\documentclass[11pt]{article}

\usepackage[preprint]{acl}

\usepackage{times}
\usepackage{latexsym}

\usepackage[T1]{fontenc}

\usepackage[utf8]{inputenc}

\usepackage{microtype}

\usepackage{inconsolata}

\usepackage{graphicx}

\usepackage{amsmath}
\usepackage{xcolor}
\usepackage{graphicx}   
\usepackage{booktabs}
\usepackage{multirow}
\usepackage{amssymb}
\usepackage{subcaption}

\title{SPAE: Spectrally Guided Autoencoder for Pretrained Visual Latents}

\author{
  \textbf{Yibin Huang\textsuperscript{1,*, $\dagger$}},
  \textbf{Jixiang Hong\textsuperscript{1,2,*, $\dagger$}},
  \textbf{Zongzhao Li\textsuperscript{1,2,*, $\dagger$}},
  \textbf{Yuhan Dai\textsuperscript{1}},\\
  \textbf{Zhibin Wang\textsuperscript{1}},
  \textbf{Chunwei Wang\textsuperscript{1}},
  \textbf{Jun Song\textsuperscript{1,$\ddagger$}},
  \textbf{Chen Wang\textsuperscript{1}},\\
  \textbf{Xiaofei Sun\textsuperscript{1}},
  \textbf{Xiaoxiao Xu\textsuperscript{1}},
  \textbf{Conghui Zhu\textsuperscript{$\ddagger$}}\\[2pt]
  \textsuperscript{1}Alibaba Group\\
  \textsuperscript{2}Gaoling School of Artificial Intelligence, Renmin University of China\\
  \small{$^*$Equal contribution.
  \quad $\dagger$This work was completed during internship.
  \quad $^\ddagger$Corresponding author.}
}

\begin{document}
\maketitle

\begin{abstract}


Latents from vision foundation models (VFMs) are semantically rich and well suited for visual understanding. Recent representation autoencoder methods such as RAE have shown that they can provide promising latent spaces for image generation. However, VFM latents remain difficult to model directly: DiT-generated latents exhibit spectral mismatch with encoder latents, especially in high-frequency components.
Our channel-wise spectral analysis further reveals that these high-frequency components are diffusely distributed across latent channels and entangled with semantic information, making the latent space difficult for DiT to model.
To address these challenges, we propose SPAE, latent adaptation framework for generation. Specifically, SPAE employs a compact bottleneck to distill stable semantic information while suppressing high-frequency components, thereby improving the alignment between DiT-generated latents and encoder latents.
In addition, we apply a channel-wise masking strategy to promote the decoupling of semantic information and high-frequency details across bottleneck channels.
Experiments show that SPAE achieves a favorable balance among visual understanding, generation quality, and reconstruction fidelity.

\end{abstract}
\section{Introduction}
\label{sec:intro}

Visual understanding and generation have long relied on different latent spaces.
Visual understanding typically builds on semantic latents extracted by VFMs, whereas generative models mainly operate in VAE latent spaces~\cite{shi2025latent}. 
Recent work, notably RAE~\cite{zheng2026diffusion}, has begun to bridge this divide by replacing the conventional VAE with semantic encoders such as DINOv2~\cite{oquab2023dinov2} and SigLIP2~\cite{tschannen2025siglip}.
Such latents encode richer semantic information, thus accelerating DiT training~\cite{peebles2023scalable}.
This shift makes VFMs a promising foundation for both visual understanding and generation~\cite{tong2026beyond}.

Despite their semantic advantages, high-dimensional latents are not ideal for direct generative modeling.
When optimized for reconstruction, these latents tend to preserve fine-grained details, which can be detrimental for generation.
Although recent studies~\cite{qiu2026hydra, yue2025uniflow} have sought to balance reconstruction and generation, a fundamental question remains: what kind of latents can preserve semantic information and reconstruction fidelity while also providing a stable space for generation?
To investigate this question, we analyze why high-dimensional semantic latents remain difficult for generative modeling.

The first obstacle is the distribution mismatch between VFM encoder latents and DiT-generated latents. During autoencoder training, the decoder receives encoder latents as input, whereas during inference it receives latents generated by DiT. 
This train–test discrepancy induces a systematic distribution shift between the two types of latents. 
We observe a mismatch mainly concentrated in high-frequency components, indicating that DiT does not model high-frequency content faithfully.

The second obstacle stems from the coupling of semantic information and high-frequency details across latent channels.
Spectral analysis further reveals that high-frequency components are broadly dispersed throughout the latent space, with their energy distributed across channels. 
As a result, stable semantic information and unstable fine-grained details remain coupled, making the latent space difficult for DiT to model~\cite{chen2025dc}.

Motivated by the aforementioned analysis, we propose SPAE, a latent adaptation framework for high-dimensional semantic latents.
Specifically, we introduce a compact bottleneck to distill stable semantic information while suppressing high-frequency components. Within the bottleneck space, we further employ channel-wise masking to promote the decoupling of semantic and detail information and thereby improve its suitability for DiT modeling.
In this way, our method better balances visual understanding, reconstruction fidelity, and generative performance.

In summary, our contributions are as follows:

\begin{itemize}

    \item \textbf{Analysis.} We identify two key obstacles to generation from high-dimensional semantic latents: a high-frequency mismatch between VFM-encoder and DiT-generated latents, and channel-wise entanglement between semantic and high-frequency components.

    \item \textbf{Method.} We propose SPAE, a latent adaptation framework with two components: a compact bottleneck that preserves semantic information while suppressing high-frequency components, and channel masking that encourages channel-wise information decoupling.
    
    \item \textbf{Experiments.} Extensive experiments show that our method achieves a balance among visual understanding, reconstruction fidelity, and generative performance.

\end{itemize}

\section{Related Work}
\label{sec:related_word}

\paragraph{Representation-based latent spaces for generation.}
Recent work has increasingly explored replacing conventional VAE latents with representation spaces derived from pretrained vision encoders. 
RAE~\cite{zheng2026diffusion} and SVG~\cite{shi2025latent} show that features from encoders such as DINO and SigLIP can serve as effective generative spaces, improving convergence and sample quality through stronger semantic structure. 
Building on this direction, subsequent studies investigate how such representations can be adapted for practical generation. 
FAE~\cite{gao2025one} maps pretrained features to lower-dimensional latents with a lightweight feature autoencoder, while RPiAE~\cite{gong2026rpiae} further preserves pretrained semantic geometry during reconstruction-oriented adaptation. 
PS-VAE~\cite{zhang2025both} argues that representation features are not directly ready for open-world generation and introduces a compact latent space with both semantic and pixel-level supervision. 
LV-RAE~\cite{liu2026improving} instead augments semantic features with missing local variations and improves robustness in high-dimensional latent spaces. 
Taken together, these works suggest that pretrained representations are promising for generation, but typically require additional adaptation before they can serve as effective generative latents.

\paragraph{Spectral structure and diffusability of latent spaces.}
Another related line of work studies what makes a latent space easier for diffusion models to learn. 
Rather than focusing on semantic priors, these works emphasize the spectral organization of latent representations. 
Recent spectral analyses~\cite{skorokhodov2025improving} show that excessive or poorly organized high-frequency components can hinder diffusion training and weaken coarse-to-fine generation. 
EQ-VAE~\cite{kouzelis2025eq} approaches this issue from the perspective of latent geometry, showing that more regular and equivariant latent spaces can improve modelability and convergence. 
UAE~\cite{fan2025prism} further provides a broader spectral interpretation, relating low-frequency components to semantic abstraction and high-frequency components to fine-grained visual detail. 
Different from these works, our method focuses on the mismatch within high-dimensional semantic representations themselves, and addresses it by retaining semantic information while suppressing unstable high-frequency components through channel-wise decoupling.

\begin{figure*}
    \centering
    \includegraphics[width=0.9\textwidth]{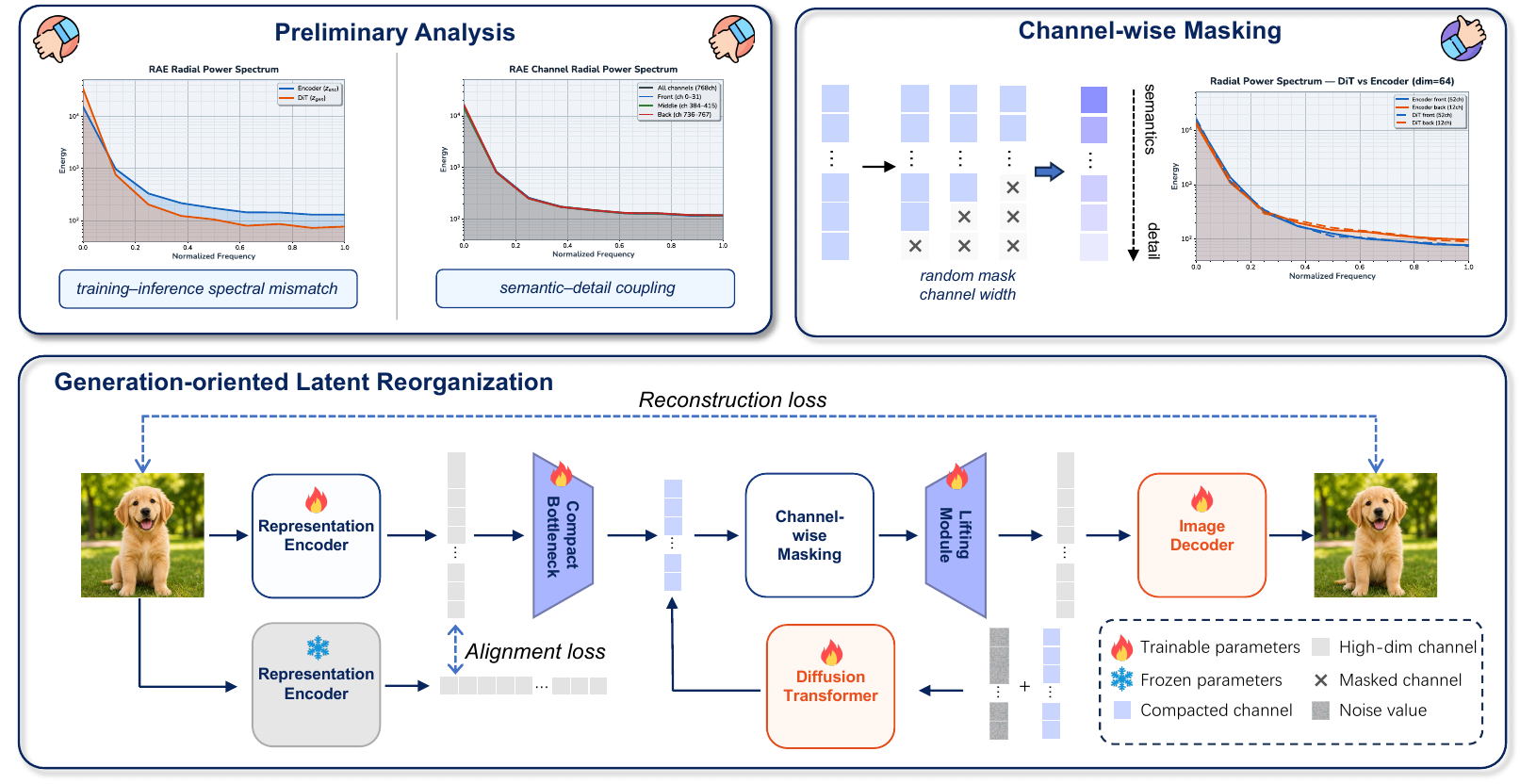}
    \caption{Overview of the SPAE. Based on spectral and channel-wise analyses of high-dimensional latents, we develop a latent adaptation framework with compact bottleneck compression and a channel-wise masking strategy.}
    \label{fig:overview}
\end{figure*}

\section{Method}
\label{sec:method}


In this section, we first analyze why high-dimensional latents are not directly suitable for generation in Section~\ref{sec:analysis_hdlatents}. 
Based on this analysis, we then introduce SPAE, a latent adaptation framework in Section~\ref{sec:method_reorg}. 
Finally, we present the stage-wise training strategy in Section~\ref{sec:training_pipeline}.

\subsection{Preliminary Analysis}
\label{sec:analysis_hdlatents}

We first define the latent space studied in this section. Given an input image $x \in \mathbb{R}^{H_0 \times W_0 \times 3}$, the representation encoder $E_{\mathrm{rep}}$ maps it to a high-dimensional latent $Z_h \in \mathbb{R}^{H \times W \times C_h}$. The decoder reconstructs the input from $Z_h$, while DiT is trained to model the same latent space for generation.

We consider two latent settings, VAEs with compact low-dimensional latents and RAEs with high-dimensional latents, and analyze them from two aspects critical for generation: spectral consistency between training and inference, and channel-wise information organization in the latent space.

\subsubsection{Training--Inference Spectral Mismatch}
\label{sec:analysis_mismatch}

The decoder is trained on latents extracted by the encoder from real images, yet at inference time it decode latents generated by DiT. 
To examine this discrepancy, we compare the spectral distributions of encoder latents and DiT-generated latents under the same class conditioning.

Specifically, we collect encoder latents and DiT-generated latents from multiple ImageNet-1K classes and samples. We then transform them into the frequency domain via a 2D Fourier transform~\cite{buchholz2022fourier}, compute the corresponding spectral energy, and radially average it to obtain radial power spectra.

\begin{figure*}[t]
    \centering
    \begin{subfigure}[t]{0.31\textwidth}
        \centering
        \includegraphics[width=\linewidth]{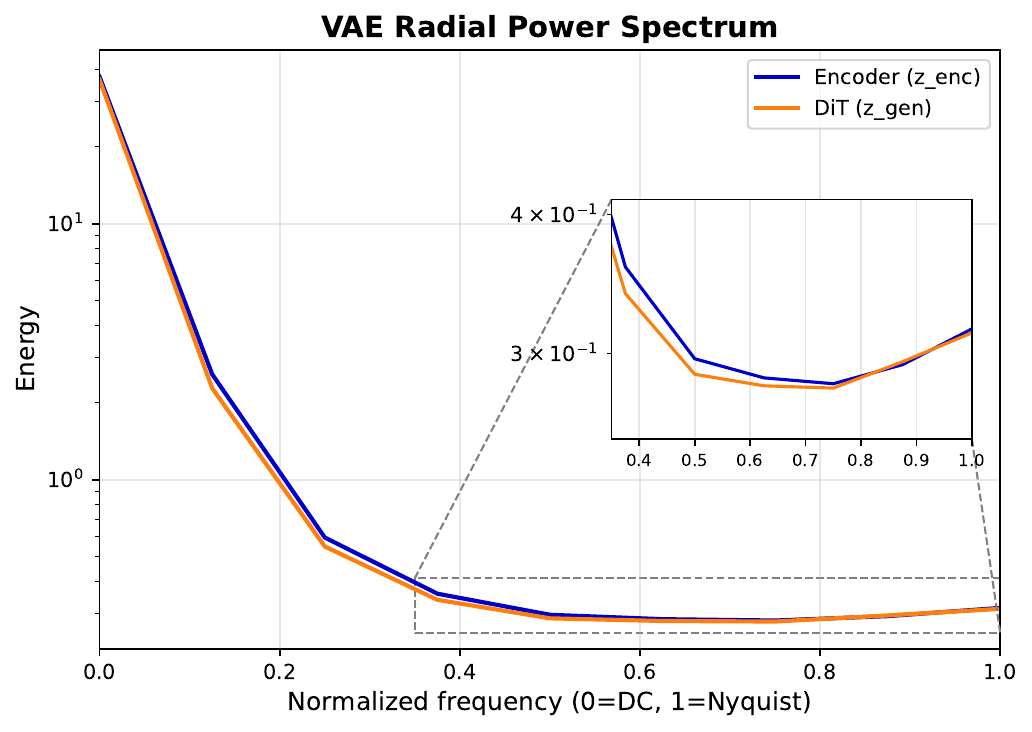}
        \caption{VAE latent space}
        \label{fig:fft_compare_a}
    \end{subfigure}
    \hfill
    \begin{subfigure}[t]{0.31\textwidth}
        \centering
        \includegraphics[width=\linewidth]{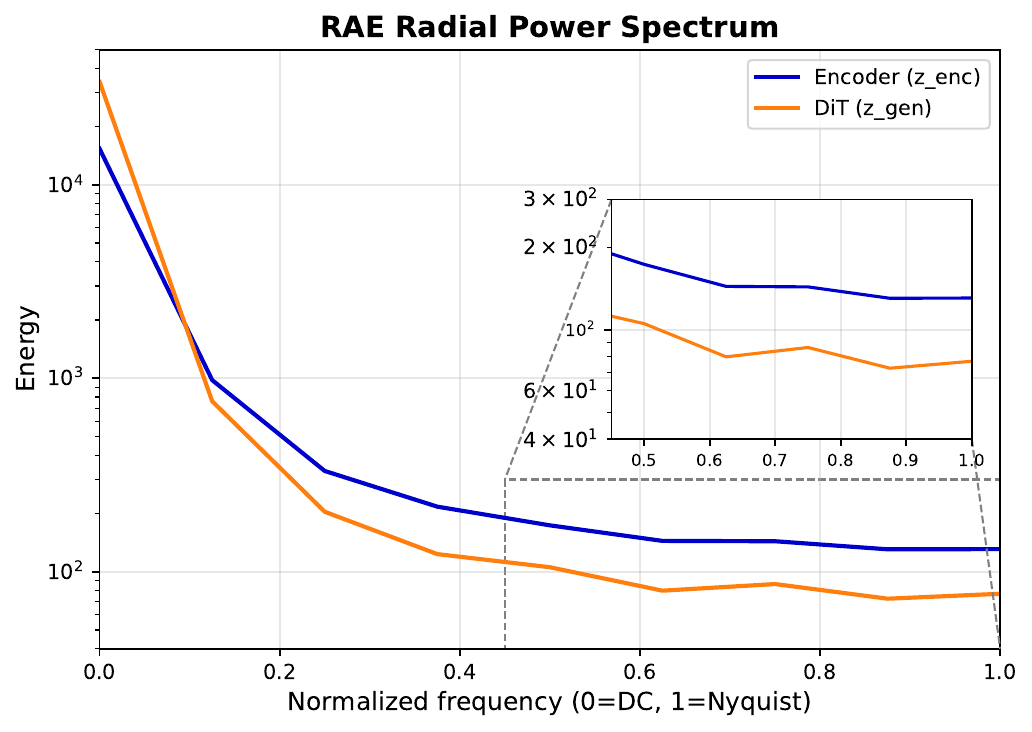}
        \caption{RAE latent space}
        \label{fig:fft_compare_b}
    \end{subfigure}
    \hfill
    \begin{subfigure}[t]{0.31\textwidth}
        \centering
        \includegraphics[width=\linewidth]{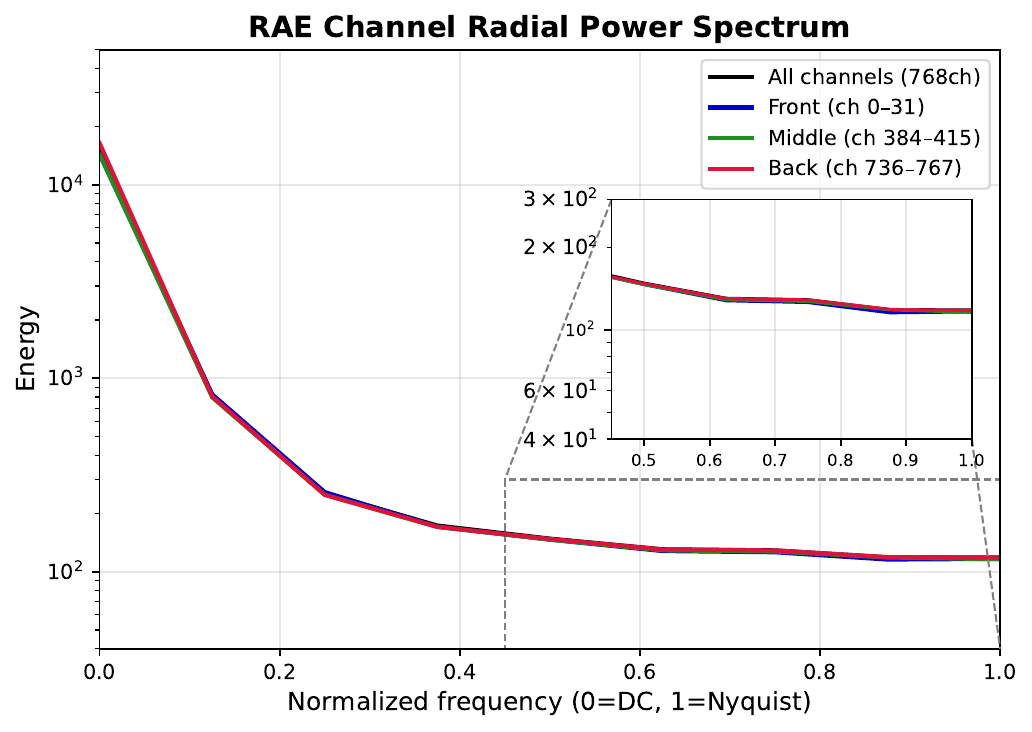}
        \caption{Channel groups in RAE latent space}
        \label{fig:fft_compare_c}
    \end{subfigure}
    \caption{Radial power spectra comparing encoder and DiT-generated latents in the VAE and RAE latent spaces, together with spectra of different channel groups in the RAE latent space. The horizontal axis denotes normalized spatial frequency, with lower frequencies on the left and higher frequencies on the right.}
    \label{fig:fft_compare}
\end{figure*}

As shown in Fig.~\ref{fig:fft_compare_a} and Fig.~\ref{fig:fft_compare_b}, for VAE latents, the radial power spectra of encoder latents and DiT-generated latents are highly similar across nearly the entire frequency range, with only minor deviations at high frequencies. For RAE latents, however, the spectral correspondence is noticeably weaker. While the spectra of RAE-encoder and DiT-generated latents remain relatively close at low frequencies, a clear gap emerges from the middle-frequency range onward, where DiT-generated latents consistently exhibit lower spectral power.

These results suggest that the training--inference spectral mismatch is less pronounced in low-dimensional latent spaces than in high-dimensional ones, indicating that reducing latent dimensionality may improve suitability for generation.

\subsubsection{Semantic--Detail Coupling}
\label{sec:analysis_channel}

Beyond spectral alignment, the generative suitability of a latent space may also depend on its channel-wise organization, as suggested by prior work~\cite{chen2025dc}.
Here we use spatial frequency as a coarse proxy for information type: lower frequencies are typically associated with global structure and semantic content, whereas higher frequencies are more related to local texture and fine details.

To assess whether low- and high-frequency components are separated along the channel dimension, we use the RAE encoder latents described above and select three channel groups of equal size from the leading, intermediate, and trailing portions of the latent.
For each group, we average over the selected channels and compute the corresponding radial power spectrum.

As shown in Fig.~\ref{fig:fft_compare_c}, the radial power spectra of different channel groups remain highly similar over the full frequency range. 
The leading, intermediate, and trailing channel groups exhibit nearly identical spectral trends and comparable magnitudes. 
This suggests that both low- and high-frequency components are broadly distributed across channels rather than concentrated in particular subsets.

These observations suggest that the high-dimensional latents lack a clear channel-wise hierarchy.
Coarse semantic information and fine details are mixed across channels, making structural signals relatively sparse and therefore more difficult to model, which may hinder generation.

\subsection{Generation-Oriented Latent Reorganization}
\label{sec:method_reorg}

The analysis suggests that high-dimension latent may be suboptimal for generation. 
Consequently, we propose SPAE, a latent adaptation framework with two components: compact bottleneck compression and structured channel-wise masking.

\subsubsection{Compact Bottleneck Compression}
\label{sec:compact_compression}

The analysis in Section~\ref{sec:analysis_mismatch} reveals a pronounced mismatch between VFM encoder latents and DiT-generated latents in the high-dimensional latent space, motivating us to compress it into a more compact space for generation.

Our design follows a simple principle: compression should serve generation, while reconstruction is handled downstream. 
Accordingly, we adopt an asymmetric architecture in which a lightweight one-layer compressor maps the native latent into a bottleneck space, and a deeper lifting module with the decoder restores reconstruction fidelity. 
This keeps the diffusion target compact, providing a more stable space for generation.

Formally, given an input image $x \in \mathbb{R}^{H_0 \times W_0 \times 3}$, we first extract the latent using the encoder $E_{\mathrm{rep}}$, compress it into a bottleneck latent via the compressor $E_c$, and then lift it back to the high-dimensional latent using the lifting module $E_u$:
\begin{equation}
Z_h = E_{\mathrm{rep}}(x), \quad
Z_b = E_c(Z_h), \quad
Z_u = E_u(Z_b)
\end{equation}

Here, $Z_h \in \mathbb{R}^{H \times W \times C_h}$ denotes the high-dimensional latent, $Z_b \in \mathbb{R}^{H \times W \times C_b}$ the compact bottleneck latent, and $Z_u \in \mathbb{R}^{H \times W \times C_h}$ the lifted latent used for decoding, where $H$ and $W$ are the latent spatial dimensions and $C_b \ll C_h$. Diffusion is carried out in the bottleneck space, and the resulting latent is then lifted by $E_u$ before decoding.

\subsubsection{Structured Channel-Wise Masking}
\label{sec:structured_masking}

As discussed in Section~\ref{sec:analysis_channel}, the native latent lacks a clear channel-wise hierarchy, which is not explicitly induced by bottleneck compression alone.

To encourage a more structured channel-wise allocation, we apply channel-wise masking during training.
By masking a suffix of channels and requiring reconstruction from the remaining ones, the model is encouraged to place low-frequency, semantically information in earlier channels and defer high-frequency details to later ones.

Specifically we randomly sample a masking width $k$ from a predefined candidate set $\mathcal{S} = \{s_1, \dots, s_M\}$, where each candidate satisfies $0 \le s_i < C_b$. For a given $k$, we preserve the first $C_b-k$ channels and mask the last $k$ consecutive channels, with $k=0$ corresponding to the unmasked case. Formally, we define a binary mask $m^{(k)} \in \{0,1\}^{C_b}$ as
\begin{equation}
m^{(k)}_i = \mathbf{1}\{i \le C_b - k\}
\end{equation}
and apply it to $Z_b$ to obtain the masked latent
\begin{equation}
\tilde{Z}_b = Z_b \odot m^{(k)}
\end{equation}
where $m^{(k)}$ is broadcast over the spatial dimensions and $\odot$ denotes element-wise multiplication.

Unlike random channel dropout, contiguous suffix masking preserves a consistent channel prefix, leading to a clearer separation between dominant semantic structure and fine-grained details in the bottleneck latent. This makes the diffusion target more structured and easier to model for generation.

\begin{table*}[!t]
    \centering
    \small
    \caption{Quantitative comparison of reconstruction performance across different autoencoders.}
    \resizebox{\textwidth}{!}{
    \begin{tabular}{lcccccccc}
    \toprule
    \multirow{2}{*}{\textbf{Tokenizer}}
    & \multicolumn{4}{c}{\textbf{ImageNet-1K}}
    & \multicolumn{4}{c}{\textbf{COCO 2017}} \\
    \cmidrule(lr){2-5} \cmidrule(lr){6-9}
    & \textbf{PSNR}$\uparrow$ & \textbf{SSIM}$\uparrow$ & \textbf{LPIPS}$\downarrow$ & \textbf{rFID}$\downarrow$
    & \textbf{PSNR}$\uparrow$ & \textbf{SSIM}$\uparrow$ & \textbf{LPIPS}$\downarrow$ & \textbf{rFID}$\downarrow$ \\
    \midrule
        SD-VAE~\cite{rombach2022high}           & 25.65 & 0.748 & 0.068 & 1.22 & 25.38 & 0.760 & 0.065 & 4.26 \\
        VA-VAE~\cite{yao2025reconstruction}           & 26.26 & 0.786 & 0.046 & 0.28 & 26.07 & 0.799 & 0.044 & 2.71 \\
        SVG~\cite{shi2025latent}              & 21.87 & 0.632 & 0.114 & 0.65 & - & - & - & - \\
        RPiAE~\cite{gong2026rpiae}            & 21.30 & 0.525 & 0.216 & 0.50 & - & - & - & - \\
        RAE~\cite{zheng2026diffusion}              & 18.05 & 0.501 & 0.247 & 0.54 & 18.36 & 0.470  & 0.254 & 12.14 \\
        PS-VAE~\cite{zhang2025both}           & 28.79 & 0.817 & 0.085 & 0.20 & - & - & - & - \\
        SPAE(SigLIP2)   & 29.64 & 0.884 & 0.038 & 0.17 & 29.37 & 0.865 & 0.059 & 2.43 \\
        SPAE(DINOv2)    & \textbf{30.28} & \textbf{0.896} & \textbf{0.033} & \textbf{0.13} & \textbf{29.56} & \textbf{0.868} & \textbf{0.054} & \textbf{2.32} \\
    \bottomrule
    \end{tabular}
    }
    \label{tab:reconstruction_result}
\end{table*}

\subsection{Training Strategy}
\label{sec:training_pipeline}

To stabilize training, we adopt a three-stage training strategy. 
The detailed design of each stage is described below.

\paragraph{Stage I: Decoder pretraining.}
We first train an image decoder $D$, instantiated as a ViT-XL architecture, while keeping the pretrained representation encoder $E_{\mathrm{rep}}$ frozen. 
At this stage, images are reconstructed directly from the high-dimensional representation. 
This stage establishes a reconstruction basis before bottleneck compression is introduced. 
The training objective is
\begin{equation}
\mathcal{L}_{\mathrm{S1}}
=
\mathcal{L}_{\mathrm{rec}}
+
\lambda_{\mathrm{perc}} \mathcal{L}_{\mathrm{perc}}
+
\lambda_{\mathrm{adv}} \mathcal{L}_{\mathrm{adv}}
\end{equation}
where $\mathcal{L}_{\mathrm{rec}}$ $\mathcal{L}_{\mathrm{perc}}$, and $\mathcal{L}_{\mathrm{adv}}$ denote the reconstruction, perceptual, and adversarial losses.

\paragraph{Stage II: Bottleneck adaptation.}
At this stage, we introduce the bottleneck modules $(E_c, E_u)$ while keeping the pretrained encoder $E_{\mathrm{rep}}$ frozen. Specifically, $E_c$ uses a single transformer block for bottleneck compression, while $E_u$ uses six transformer blocks to map the bottleneck latent back to the decoder latent space. We jointly optimize $E_c$ and $E_u$ to reconstruct images from the bottleneck latent, thereby adapting the model for subsequent end-to-end finetuning. The objective remains the same as in Stage I.

\paragraph{Stage III: End-to-end finetuning.}
In the final stage, we jointly finetune $E_{\mathrm{rep}}$, $E_c$, $E_u$, and $D$ end-to-end, while applying channel-wise masking in the bottleneck space.
Once the encoder $E_{\mathrm{rep}}$ is unfrozen, optimizing for reconstruction may cause its latent space to drift away from the pretrained one. 
To regularize such drift, we keep a frozen copy of the pretrained encoder, denoted by $E_{\mathrm{rep}}^{\mathrm{frozen}}$, and use it to impose a semantic alignment loss:
\begin{equation}
\mathcal{L}_{\mathrm{align}}
=
\left\|
E_{\mathrm{rep}}(x)
-
E_{\mathrm{rep}}^{\mathrm{frozen}}(x)
\right\|_2^2
\end{equation}
The overall objective in this stage is
\begin{equation}
\mathcal{L}_{\mathrm{S3}}
=
\mathcal{L}_{\mathrm{rec}}
+
\lambda_{\mathrm{perc}} \mathcal{L}_{\mathrm{perc}}
+
\lambda_{\mathrm{adv}} \mathcal{L}_{\mathrm{adv}}
+
\lambda_{\mathrm{align}} \mathcal{L}_{\mathrm{align}}
\end{equation}

\section{Experiments}
\label{sec:experiment}

\begin{figure*}[t]
    \centering
    \includegraphics[width=0.9\textwidth]{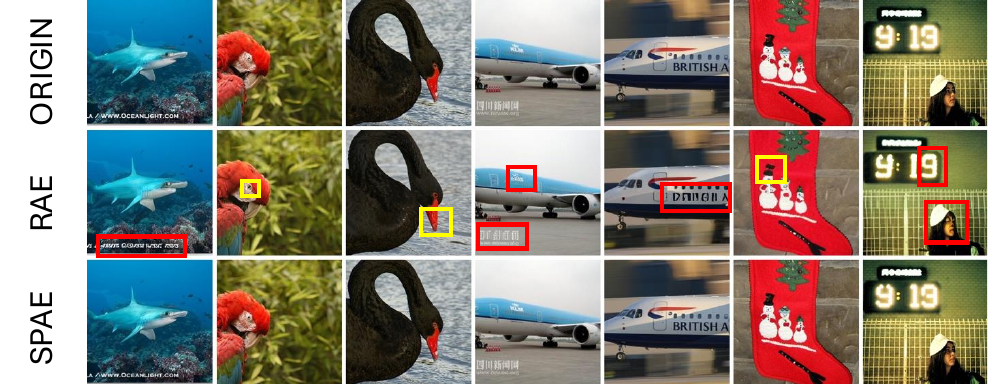}
    \caption{Reconstruction comparison between SPAE and RAE. Boxes indicate local regions for visual comparison.}
    \label{fig:recon}
\end{figure*}

\subsection{Experimental Settings}

\paragraph{Training.}
We train all models on ImageNet-1K~\cite{russakovsky2015imagenet} at $256\times256$ resolution.  During training, channel-wise masking is applied by masking the trailing $w$ channels with $w \in \{0,12,16,24,32\}$, sampled with probabilities $\{0.6, 0.15, 0.1, 0.1, 0.05\}$, respectively. 
For generation, we use $\text{DiT}^{\text{DH}}$~\cite{zheng2026diffusion} as the diffusion backbone.
More implementation details are provided in Appendix~\ref{sec:appendix_hparams}.

\paragraph{Evaluation.}
We evaluate SPAE from four perspectives: reconstruction, generation, linear-probe image classification~\cite{alain2016understanding}, and downstream multimodal understanding benchmarks. 
For reconstruction, we evaluate on the ImageNet-1K~\cite{imagenet15russakovsky} and COCO 2017~\cite{lin2014microsoft} validation sets at a resolution of $256\times256$, and report PSNR, SSIM, LPIPS~\cite{zhang2018unreasonable} and rFID. 
For generation, we consider class-conditional image generation, reporting gFID~\cite{heusel2017gans}, IS~\cite{salimans2016improved}, Precision, and Recall. For linear-probe classification, we evaluate on CIFAR-100~\cite{krizhevsky2009learning}, Food101~\cite{bossard14}, ImageNet-1K~\cite{imagenet15russakovsky}, and SUN397~\cite{5539970}. 
For downstream multimodal understanding tasks, we use the Qwen3-8B~\cite{qwen3technicalreport} as backbone and evaluate on MME~\cite{fu2026mme}, MMB~\cite{liu2024mmbench}, POPE~\cite{li2023evaluating}, TextVQA~\cite{singh2019towards}, and SQA~\cite{lu2022learn}.

\subsection{Image Reconstruction}

\paragraph{Quantitative results.}

As shown in Table~\ref{tab:reconstruction_result}, SPAE achieves strong reconstruction performance on both ImageNet-1K and COCO 2017. On ImageNet-1K, SPAE(DINOv2) achieves the best overall results, with PSNR (30.28), SSIM (0.896), LPIPS (0.033), and rFID (0.13), outperforming all compared methods across all metrics.
SPAE(SigLIP2) also performs favorably, surpassing VA-VAE by 3.38 in PSNR and 0.098 in SSIM, while reducing LPIPS from 0.046 to 0.038 and rFID from 0.28 to 0.17. 
On COCO 2017, SPAE remains consistently strong, achieving the best overall performance.

Furthermore, Fig.~\ref{fig:recon} shows that SPAE reconstructs structures and textures more faithfully than RAE, especially in the boxed regions.

\subsection{Class-Conditional Generation}

\paragraph{Quantitative results.}

As shown in Table~\ref{tab:generation_result}, SPAE performs favorably among latent diffusion methods under both unguided and guided generation. In the unguided setting, it obtains the best gFID of 1.47, slightly better than RAE (1.51) and clearly better than VA-VAE (2.17) and RPiAE (2.25). With guidance, SPAE also achieves the best gFID of 1.12 and the highest recall of 0.70, compared with RAE (1.13, 0.67) and VA-VAE (1.35, 0.65). While its IS is not the highest, it remains competitive. Taken together, these results indicate that SPAE achieves strong generation performance.

\begin{figure*}[t]
    \centering
    \includegraphics[width=0.9 \textwidth]{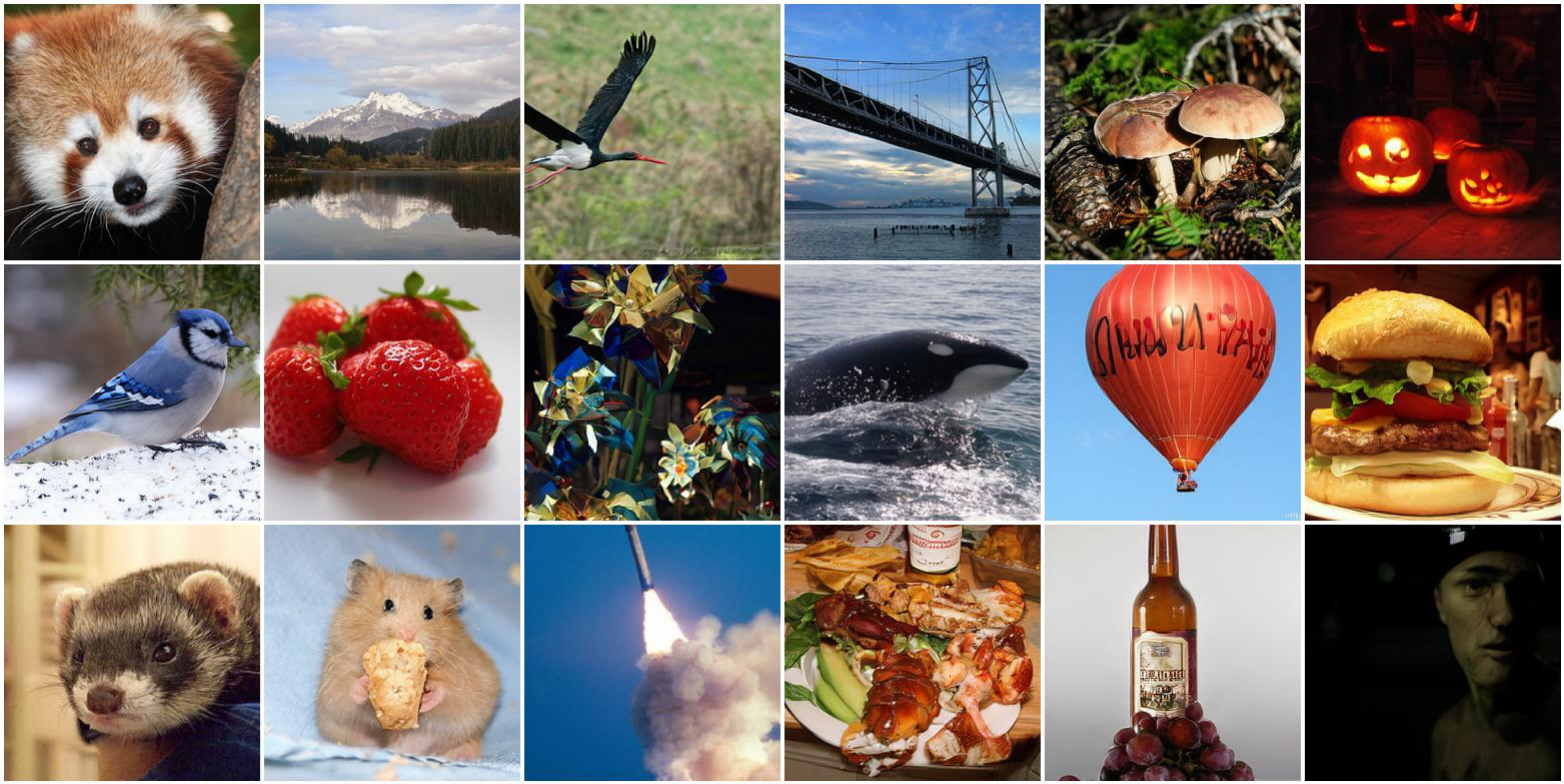}
    \caption{Class-conditional generation results without guidance. Random samples generated by SPAE.}
    \label{fig:gen}
\end{figure*}

Furthermore, Fig.~\ref{fig:gen} shows random class-conditional samples generated by SPAE on ImageNet-1K, which are visually coherent and diverse across a wide range of categories.

\begin{table*}[!t]
  \centering
  \caption{Class-conditional generation performance on ImageNet 256×256.}
  \resizebox{\textwidth}{!}{
  \begin{tabular}{lcccccccccc}
  \toprule
      \multirow{2}{*}{\textbf{Method}} 
      & \multirow{2}{*}{\textbf{\#Params}} & \multirow{2}{*}{\textbf{Epochs}}
      & \multicolumn{4}{c}{\textbf{Generation@256 w/o guidance}}
      & \multicolumn{4}{c}{\textbf{Generation@256 w/ guidance}} \\
      \cmidrule(lr){4-7} \cmidrule(lr){8-11}
      & & & \textbf{gFID}$\downarrow$ & \textbf{IS}$\uparrow$ & \textbf{Prec.}$\uparrow$ & \textbf{Rec.}$\uparrow$
       & \textbf{gFID}$\downarrow$ & \textbf{IS}$\uparrow$ & \textbf{Prec.}$\uparrow$ & \textbf{Rec.}$\uparrow$ \\
  \midrule
      \multicolumn{11}{l}{\textit{\textbf{Latent Diffusion}}} \\
  \midrule
      MaskDiT~\cite{Zheng2024MaskDiT} & 675M & 1600 & 5.69 & 177.9 & 0.74 & 0.60 & 2.28 & 276.6 & 0.80 & 0.61 \\
      DiT~\cite{Peebles2022DiT} & 675M & 1400 & 9.62 & 121.5 & 0.67 & 0.67 & 2.27 & 278.2 & \textbf{0.83} & 0.57 \\
      SiT~\cite{ma2024sit} & 675M & 1400 & 8.61 & 131.7 & 0.68 & 0.67 & 2.06 & 270.3 & 0.82 & 0.59 \\
      FasterDiT~\cite{yao2024fasterdit} & 675M & 400 & 7.91 & 131.3 & 0.67 & \textbf{0.69} & 2.03 & 264.0 & 0.81 & 0.60 \\
      VA-VAE~\cite{yao2025reconstruction} & 675M & 800 & 2.17 & 205.6 & 0.77 & 0.65 & 1.35 & \textbf{295.3} & 0.79 & 0.65 \\
      SVG~\cite{shi2025latent} & 675M & 1400 & 3.36 & 181.2 & - & - & 1.92 & 264.9 & - & - \\
      RPiAE~\cite{gong2026rpiae} & 675M & 80 & 2.25 & 208.7 & \textbf{0.81} & 0.60 & 1.51 & 225.9 & 0.79 & 0.65 \\ 
      RAE~\cite{zheng2026diffusion} & 839M & 800 & 1.51 & \textbf{242.9} & 0.79 & 0.63 & 1.13 & 262.6 & 0.78 & 0.67 \\
      SPAE & 839M & 800 & \textbf{1.47} & 240.6 & 0.77 & 0.63 & \textbf{1.12} & 280.1 & 0.75 & \textbf{0.70} \\
  \bottomrule
  \end{tabular}
  }
  \label{tab:generation_result}
\end{table*}

\begin{table*}[t]
\centering
\caption{Visual understanding results. Left: linear probe accuracy (\%). Right: downstream multimodal understanding performance. ``+ SPAE'' denotes our method built on the pretrained encoder in the preceding row.}
\label{tab:semantic}

\begin{subtable}[t]{0.44\textwidth}
\centering
\small
\caption{Linear probe accuracy (\%)}
\label{tab:linear_probe}
{
\setlength{\tabcolsep}{3.5pt}
\begin{tabular}{lcccc}
\toprule
\textbf{Model} & \textbf{CIFAR-100} & \textbf{Food101} & \textbf{ImageNet-1K} & \textbf{SUN397} \\
\midrule
DINOv2        & 90.20 & 91.54 & 80.33 & 76.31 \\
\quad + SPAE  & 89.84 & 90.63 & 80.13 & 76.11 \\
SigLIP2       & 85.77 & 93.17 & 79.77 & 78.19 \\
\quad + SPAE  & 86.21 & 92.50 & 80.07 & 77.41 \\
\bottomrule
\end{tabular}
}
\end{subtable}
\hfill
\begin{subtable}[t]{0.48\textwidth}
\centering
\small
\caption{Downstream multimodal understanding}
\label{tab:visual_understanding}
{
\setlength{\tabcolsep}{4pt}
\begin{tabular}{lccccc}
\toprule
\textbf{Model} & \textbf{MME} & \textbf{MMB} & \textbf{POPE} & \textbf{TextVQA} & \textbf{SQA} \\
\midrule
\multicolumn{6}{l}{\textit{\textbf{Qwen3-8B-Base}}} \\
\midrule
SigLIP2       & 1817 & 65.3 & 85.5 & 29.09 & 77.79 \\
\quad + SPAE  & 1896 & 67.3 & 86.1 & 35.75 & 81.68 \\
\bottomrule
\end{tabular}
}
\end{subtable}
\end{table*}

\subsection{Visual Understanding}

\paragraph{Linear probe results.}

As shown in Table~\ref{tab:linear_probe}, linear-probe accuracy remains comparable to that of the pretrained models across all benchmarks for both DINOv2 and SigLIP2. 
This suggests that SPAE largely preserves the semantic information of the original encoders.

\paragraph{Downstream understanding tasks.}

Table~\ref{tab:visual_understanding} shows that SPAE consistently improves over the original SigLIP2 encoder in the Qwen3-8B setting across all evaluated benchmarks. 
Gains are observed on both general multimodal benchmarks and reasoning-oriented tasks, indicating that the latent retains strong semantic utility beyond generation.
Overall, these results suggest that SPAE achieves a favorable balance between generative modeling and downstream multimodal understanding.

\begin{figure}[t]
    \centering
    \includegraphics[width=0.9 \linewidth]{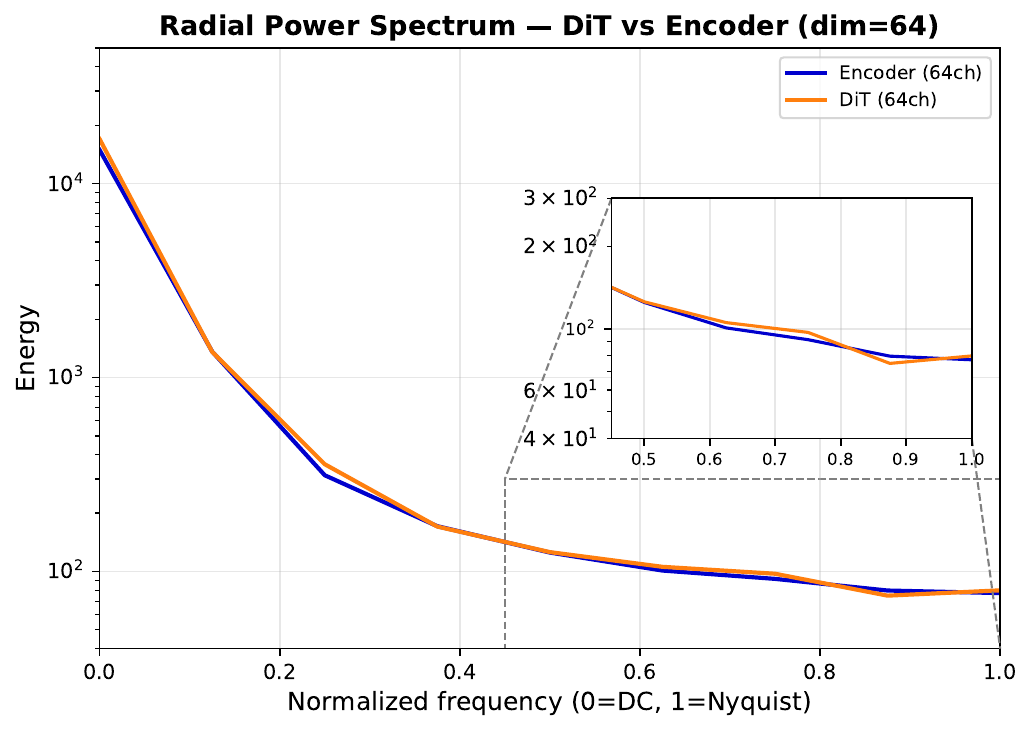}
    \caption{Radial power spectrum comparison between encoder and DiT latents in the 64-dimensional bottleneck space over all latent channels.}
    \label{fig:spectral_analysis}
\end{figure}

\begin{figure*}[t]
    \centering
    \begin{subfigure}[t]{0.24\textwidth}
        \centering
        \includegraphics[width=\linewidth]{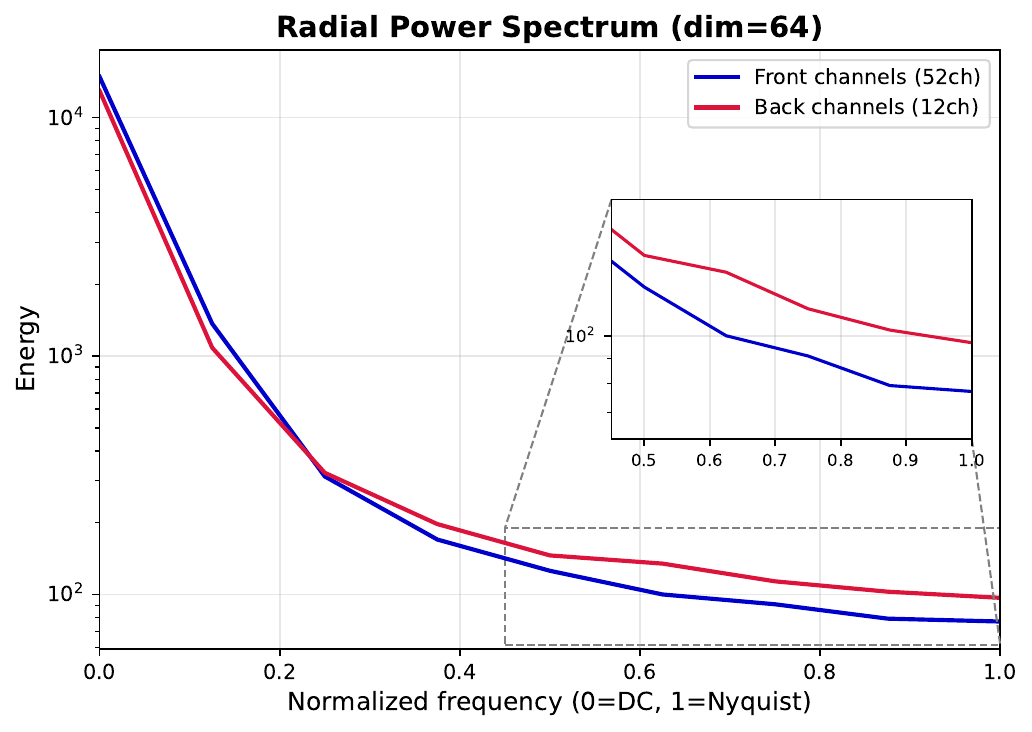}
        \caption{Channel split with $w=12$.}
        \label{fig:channel_fft_a}
    \end{subfigure}
    \hfill
    \begin{subfigure}[t]{0.24\textwidth}
        \centering
        \includegraphics[width=\linewidth]{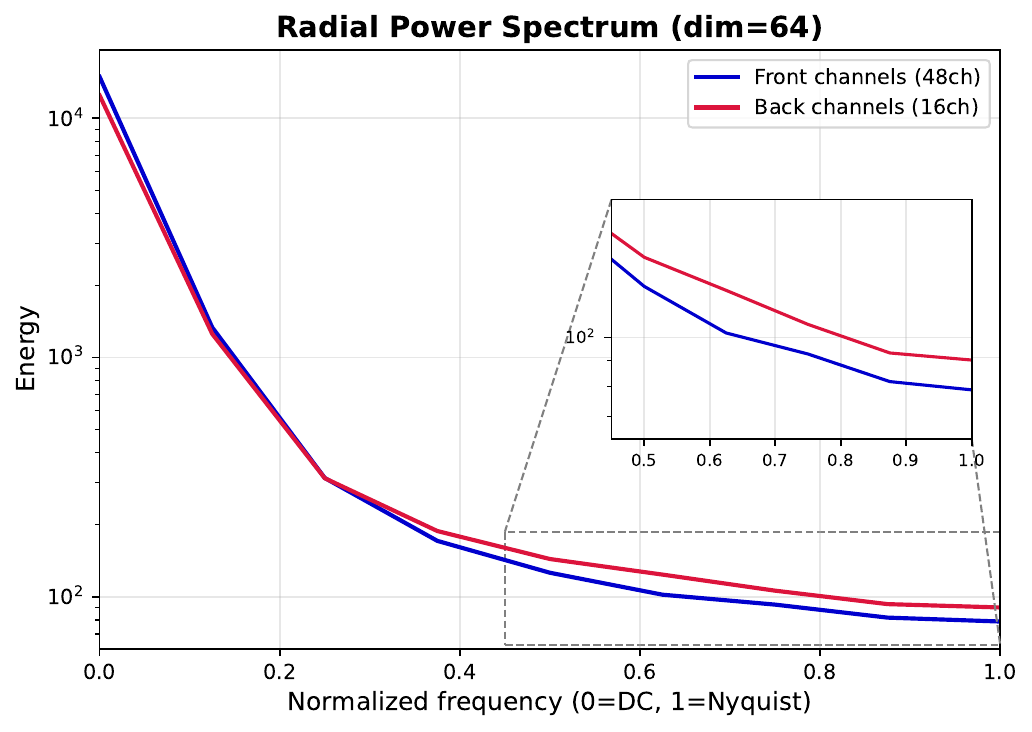}
        \caption{Channel split with $w=16$.}
        \label{fig:channel_fft_b}
    \end{subfigure}
    \hfill
    \begin{subfigure}[t]{0.24\textwidth}
        \centering
        \includegraphics[width=\linewidth]{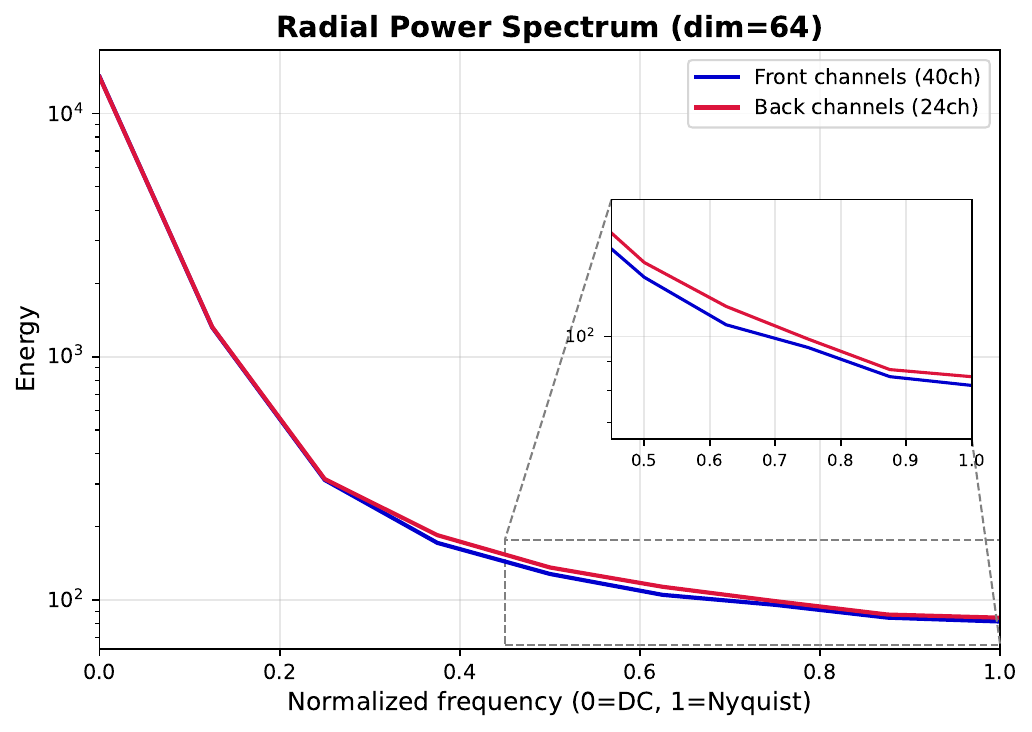}
        \caption{Channel split with $w=24$.}
        \label{fig:channel_fft_c}
    \end{subfigure}
    \hfill
    \begin{subfigure}[t]{0.24\textwidth}
        \centering
        \includegraphics[width=\linewidth]{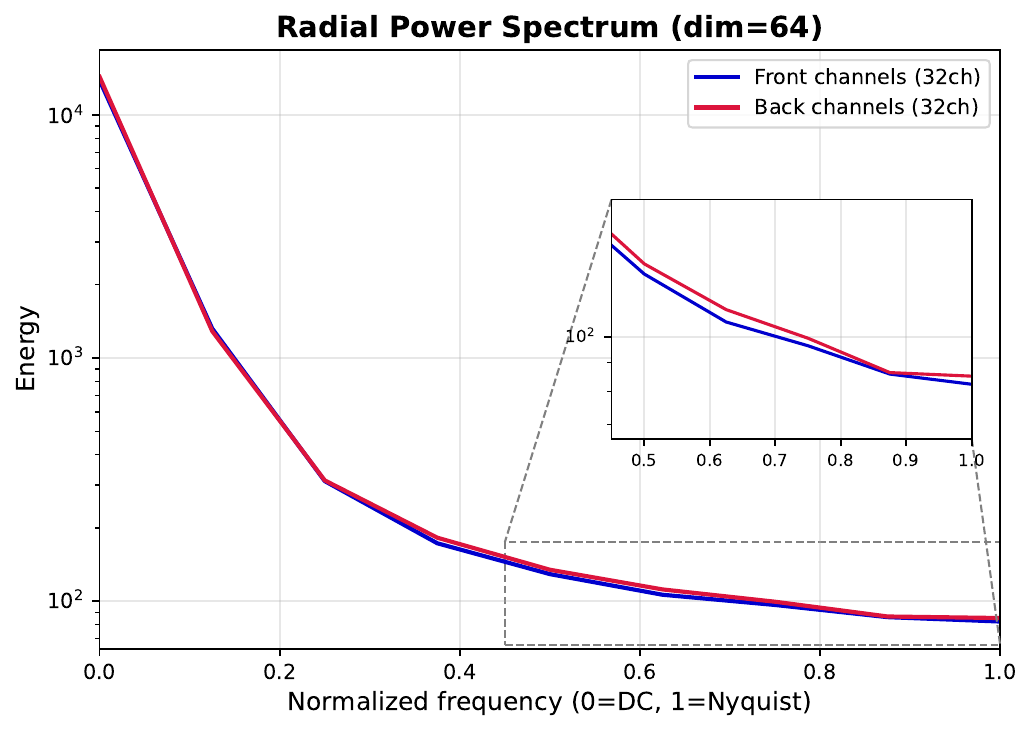}
        \caption{Channel split with $w=32$.}
        \label{fig:channel_fft_d}
    \end{subfigure}
    \caption{Channel-wise radial power spectra of 64-dimensional bottleneck latents for different channel splits. The latent channels are divided into front and back groups, $w$ denotes the number of trailing channels in the back group.}
    \label{fig:channel_fft}
\end{figure*}

\subsection{Analysis and Ablation Studies}

\begin{table*}[!t]
    \centering
    \caption{Ablation results on ImageNet-1K at 10 epochs. The effects of bottleneck dimension and masking strategy are analyzed for both reconstruction and class-conditional generation.}

    \begin{subtable}[t]{0.48\textwidth}
        \centering
        \small
        \caption{Bottleneck dimension}
        \label{tab:ablation_dim}
        \setlength{\tabcolsep}{5pt}
        \begin{tabular}{ccccc}
        \toprule
        \multirow{2}{*}{\textbf{Dimension}} & \multicolumn{2}{c}{\textbf{Reconstruction}} & \multicolumn{2}{c}{\textbf{Generation}} \\
        \cmidrule(lr){2-3} \cmidrule(lr){4-5}
        & \textbf{PSNR}$\uparrow$ & \textbf{rFID}$\downarrow$ & \textbf{gFID}$\downarrow$ & \textbf{IS}$\uparrow$ \\
        \midrule
        32  & 27.31 & 0.42 & \textbf{4.82} & 150.6 \\
        64  & 30.28 & 0.13 & 5.00 & \textbf{151.2} \\
        96  & 30.97 & 0.11 & 5.93 & 120.3 \\
        128 & \textbf{31.35} & \textbf{0.10} & 7.18 & 106.9 \\
        \bottomrule
        \end{tabular}
    \end{subtable}
    \hfill
    \begin{subtable}[t]{0.48\textwidth}
        \centering
        \small
        \caption{Masking strategy}
        \label{tab:ablation_mask}
        \setlength{\tabcolsep}{5pt}
        \begin{tabular}{ccccc}
        \toprule
        \multirow{2}{*}{\textbf{Strategy}} & \multicolumn{2}{c}{\textbf{Reconstruction}} & \multicolumn{2}{c}{\textbf{Generation}} \\
        \cmidrule(lr){2-3} \cmidrule(lr){4-5}
        & \textbf{PSNR}$\uparrow$ & \textbf{rFID}$\downarrow$ & \textbf{gFID}$\downarrow$ & \textbf{IS}$\uparrow$ \\
        \midrule
        Unmask       & \textbf{30.96} & \textbf{0.12} & 8.30 & 97.5 \\
        Token mask   & 30.33 & 0.13 & 7.54 & 102.9 \\
        Channel mask & 30.28 & 0.13 & \textbf{5.00} & \textbf{151.2} \\
        \bottomrule
        \end{tabular}
    \end{subtable}

    \label{tab:ablation_main}
\end{table*}

\paragraph{Effect of bottleneck compression.}

To study the role of bottleneck compression, we analyze it from both spectral and quantitative perspectives.

We first compare the frequency spectra of encoder and DiT-generated latents in high-dimensional space and in 64-dimensional space. As shown in Fig.~\ref{fig:fft_compare_b} and Fig.~\ref{fig:spectral_analysis}, the spectral mismatch is substantially reduced after compression. In the bottleneck space, the generated spectra follow the encoder spectra more closely, especially in the high-frequency range, suggesting that bottleneck compression suppresses latent components that are difficult for DiT to model.

We further vary the bottleneck dimension to examine the trade-off between reconstruction and generation. Table~\ref{tab:ablation_dim} shows that reconstruction improves steadily with wider bottlenecks: PSNR rises from 27.31 at dimension 32 to 31.35 at dimension 128. In contrast, generative quality deteriorates as the bottleneck becomes wider, with the best gFID achieved at dimension 32 and progressively worse results at larger dimensions.

Overall, these results show that bottleneck compression makes the latent space more amenable to diffusion modeling, while its dimensionality controls the balance between reconstruction fidelity and generative quality. We therefore adopt dimension 64 as the default setting, which provides a favorable trade-off between the two.

\paragraph{Effect of channel-wise masking.}

We next investigate whether channel-wise masking encourages a more structured bottleneck latent and how such structure affects generation. Specifically, we analyze the spectral characteristics of different channel groups in the 64-dimensional bottleneck latent and compare three masking strategies: no masking, the proposed channel-wise masking, and a token-masking baseline matched in both masking ratio and masking probability schedule.

Fig.~\ref{fig:channel_fft} reveals a consistent frequency separation across all channel splits: the front channels preserve relatively more low-frequency energy, whereas the back channels retain relatively more high-frequency components. 
This trend becomes increasingly pronounced as the back group becomes smaller, and is most evident at $w=12$, which corresponds to the highest masking probability used during training.
These observations suggest that channel-wise masking encourages an ordered organization of the bottleneck latent, in which coarse semantic information is concentrated in earlier channels, while finer visual details are encoded in later channels.

Table~\ref{tab:ablation_mask} shows that masking affects reconstruction only marginally, while having a more noticeable effect on generation.
Compared with the unmasked setting, channel-wise masking causes only a slight drop in reconstruction quality, with PSNR decreasing from 30.96 to 30.28 and rFID remaining at 0.13. 
In contrast, it substantially improves generation, reducing gFID from 8.30 to 5.00 and increasing IS from 97.5 to 151.2. 
 channel-wise masking induces a latent structure that is especially beneficial for generation, while still retains most of the information necessary for reconstruction.

By comparison, token masking yields only limited gains, reaching gFID (7.54) and IS (102.9), and still underperforms channel-wise masking by a clear margin. 
This indicates that the benefit comes not simply from masking itself, but from a masking pattern that is better aligned with the structure of the bottleneck latent.

\section{Conclusion}
\label{sec:conclusion}

In this work, we study how to make visual latents more suitable for visual understanding, reconstruction, and generative modeling. 
We show that, high dimension latents are not naturally suited to generation: unstable high-frequency components lead to a mismatch between training and inference, while semantic content entangled with fine details across channels. 
Consequently, we propose SPAE, a generation-oriented latent adaption framework based on compact bottleneck compression and structured channel-wise masking. 
SPAE improves the generative suitability while preserving their strengths in reconstruction and semantic understanding. 
We hope this work serves as a useful step toward visual encoders that can better support unified multimodal models.

\section{Limitation}
\label{sec:limitaion}

While SPAE achieves strong results in both generation and understanding, several aspects remain worthy of further study. First, our use of a fixed training resolution may limit the modeling of fine-grained local information, especially in text-rich regions, thereby affecting performance on tasks such as TextVQA. Second, we use a fixed channel-wise masking schedule throughout all experiments. Although this design works well in our current setting, more adaptive masking strategies that depend on the data, model scale, or target task may further improve the learned latent structure.

\section*{Ethical Considerations}
All authors affirm that this work complies with the ACL Code of Ethics. The authors declare no conflicts of interest related to this research. All code, models, and datasets used in this study are publicly available and released under their respective open-source licenses (e.g., Apache-2.0). To further support transparency and reproducibility, we also document the remaining limitations of this work, along with all prompt templates and evaluation details, in the Appendix.

\bibliography{latex/acl_latex}

\appendix

\appendix
\section{Appendix}
\label{sec:appendix}

\subsection{Training details}
\label{sec:appendix_hparams}

We use pretrained vision encoders as latent extractors, including DINOv2-Base and SigLIP2-Base. 
Unless otherwise specified, all main results are reported using DINOv2-Base. 
The input resolution is matched to the corresponding encoder backbone, i.e., $224 \times 224$ for DINOv2-Base and $256 \times 256$ for SigLIP2-Base.

\paragraph{Autoencoder training.}
We train the autoencoder in three stages. 
In Stage 1, we train the decoder to reconstruct images from frozen encoder latents without applying bottleneck compression. 
This stage allows the decoder to adapt to the pretrained latent space. 
In Stage 2, we introduce a 64-dimensional bottleneck and train only the compressor, while keeping both the encoder and decoder fixed. 
In Stage 3, we jointly finetune all mudules under latent masking. 
The stage-wise hyperparameters are summarized in Table~\ref{tab:ae_hparams}.

In Stage 3, we use the structured channel-wise masking strategy described in Section~\ref{sec:structured_masking}. 
The masking distribution is given in Table~\ref{tab:masking_hparams}. 
At each training step, we sample a masking level from this distribution and mask the corresponding number of channels in the 64-dimensional bottleneck. 
As an ablation baseline, we also consider token masking. 
To ensure a fair comparison, token masking follows the same masking probability distribution as channel masking, but is applied over spatial tokens rather than channels. 
Specifically, a subset of spatial tokens is randomly selected and replaced with a learnable mask token, while keeping the overall masking schedule aligned.

\paragraph{DiT training.}
After training the autoencoder, we freeze it and train the latent generator in the normalized bottleneck latent space. 
We use \texttt{DiTwDDTHead} with hidden size $[1152, 2048]$, depth $[28, 2]$, and 16 attention heads per block. 
The model is trained on ImageNet-1K with 1000 classes and a class dropout ratio of 0.1. 
We adopt flow matching with a linear path and velocity prediction as the training objective. 
At inference time, we use an ODE sampler with Euler discretization and 50 sampling steps. 
Training runs for 1400 epochs with an effective batch size of 1024 using AdamW, a learning rate of $2 \times 10^{-4}$, gradient clipping of 1.0, and an EMA decay of 0.9995 in fp32 precision. 
The learning rate follows a linear decay schedule with a 40-epoch warmup and decays until epoch 800.

\paragraph{Multimodal understanding training.}
For downstream multimodal understanding experiments, we adopt a LLaVA-style architecture consisting of a frozen vision encoder, a two-layer MLP projector with GELU activations, and Qwen3-8B as the language backbone. 
Training is performed in two stages. 
In Stage 1, we pretrain the projector on the ShareGPT4V 1.3M caption dataset while keeping both the vision encoder and the language model frozen. 
In Stage 2, we finetune the projector together with the language model on the LLaVA-665K instruction-following dataset, while still freezing the vision encoder. 
The overall setup is summarized in Table~\ref{tab:mm_hparams}. 
Unless otherwise stated, the SigLIP2 baseline and SPAE use the same training hyperparameters and differ only in the initialization of the vision encoder.

\begin{table*}[t]
\centering
\small
\caption{Training hyperparameters for the three-stage autoencoder training procedure.}
\label{tab:ae_hparams}
\begin{tabular}{lccc}
\toprule
 & Stage 1 & Stage 2 & Stage 3 \\
\midrule
trainable modules & decoder & compressor & encoder + compressor + decoder \\
epochs & 16 & 2 & 10 \\
batch size & 512 & 512 & 512 \\
optimizer & AdamW & AdamW & AdamW \\
learning rate & $2 \times 10^{-4}$ & $2 \times 10^{-4}$ & $2 \times 10^{-5}$ \\
loss & $\ell_1 + \mathrm{LPIPS} + \mathrm{GAN}$ & $\ell_1 + \mathrm{LPIPS}$ & $\ell_1 + \mathrm{LPIPS} + \mathrm{GAN}$ \\
disc start & epoch 8 & -- & epoch 5 \\
disc lr & $2 \times 10^{-4}$ & -- & $5 \times 10^{-5}$ \\
\bottomrule
\end{tabular}
\end{table*}

\begin{table*}[t]
\centering
\small
\caption{Masking configurations used in Stage 3. Channel masking is used in the main experiments. Token masking serves as an ablation baseline and follows the same masking probability distribution, but randomly replaces the selected spatial tokens with a learnable mask token.}
\label{tab:masking_hparams}
\begin{minipage}[t]{0.47\textwidth}
\centering
\textbf{Channel masking}
\vspace{0.3em}

\begin{tabular}{lccccc}
\toprule
probability & 0.60 & 0.15 & 0.10 & 0.10 & 0.05 \\
\midrule
masked channels & 0 & 12 & 16 & 24 & 32 \\
retained channels & 64 & 52 & 48 & 40 & 32 \\
\bottomrule
\end{tabular}
\end{minipage}
\hfill
\begin{minipage}[t]{0.47\textwidth}
\centering
\textbf{Token masking}
\vspace{0.3em}

\begin{tabular}{lccccc}
\toprule
probability & 0.60 & 0.15 & 0.10 & 0.10 & 0.05 \\
\midrule
masked tokens & 0 & 48 & 64 & 96 & 128 \\
retained tokens & 256 & 208 & 192 & 160 & 128 \\
\bottomrule
\end{tabular}
\end{minipage}
\end{table*}

\begin{table}[t]
\centering
\small
\caption{Multimodal understanding training setup.}
\label{tab:mm_hparams}
\begin{tabular}{lcc}
\toprule
parameter & Stage 1 & Stage 2 \\
\midrule
objective & projector pretraining & instruction tuning \\
training data & ShareGPT4V-1.3M & LLaVA-665K \\
trainable modules & projector & projector + LLM \\
learning rate & $1 \times 10^{-3}$ & $2 \times 10^{-5}$ \\
optimizer & AdamW & AdamW \\
scheduler & cosine & cosine \\
warmup ratio & 0.03 & 0.03 \\
max length & 4096 & 2048 \\
\bottomrule
\end{tabular}
\end{table}

\subsection{Evaluation and additional results}
\label{sec:appendix_eval_more}

\paragraph{Evaluation protocol.}
All experiments are conducted on publicly available datasets released through Hugging Face. 
No private, confidential, or personally restricted data are used in this work. 
Each experiment is repeated with three different random seeds, and we report the best result across the three runs.

\paragraph{Linear-probe evaluation.}
We follow the standard linear-probe protocol, in which a single linear classifier is trained on top of frozen encoder latents. 
Specifically, we train a linear layer using AdamW for 10 epochs with a learning rate of 0.1 and a cosine decay schedule without warmup. 
The batch size is 64, the loss is cross-entropy, and the input features are L2-normalized. 
We evaluate on CIFAR-10, CIFAR-100, Food-101, Caltech-101, SUN397, and ImageNet-1K, using the standard train/test splits, except for ImageNet-1K, where we use the train/validation split.

\begin{figure*}[t]
    \centering
    \begin{subfigure}[t]{0.47\textwidth}
        \centering
        \includegraphics[width=\linewidth]{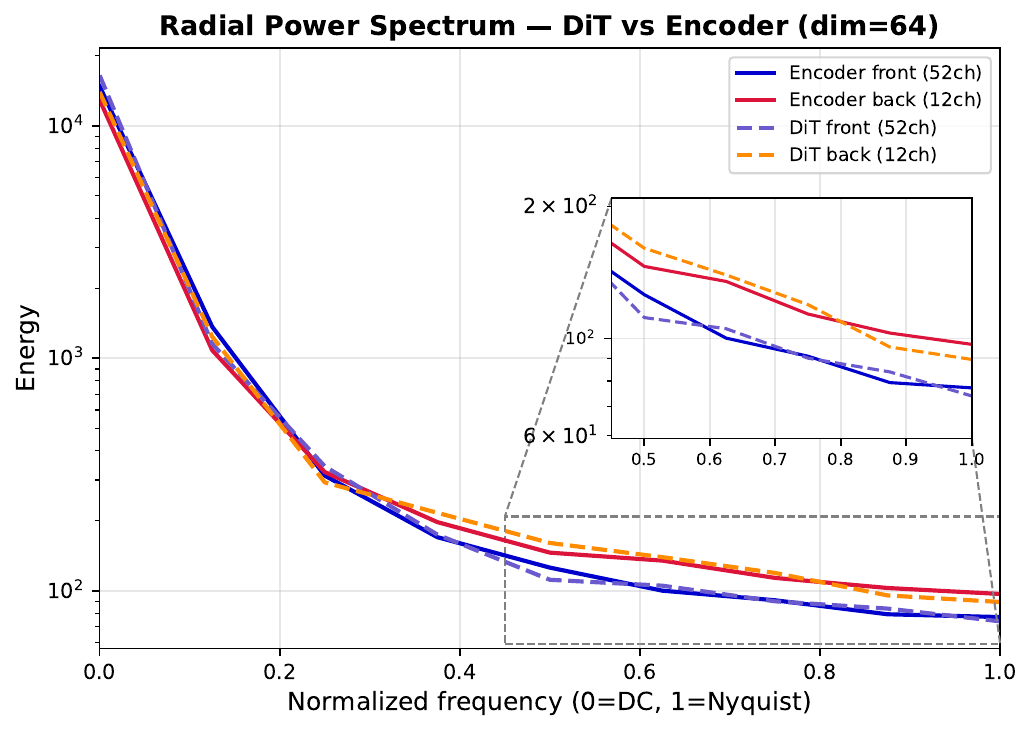}
        \caption{Channel split with $w=12$.}
        \label{fig:appendix_ddt_encoder_a}
    \end{subfigure}
    \hfill
    \begin{subfigure}[t]{0.47\textwidth}
        \centering
        \includegraphics[width=\linewidth]{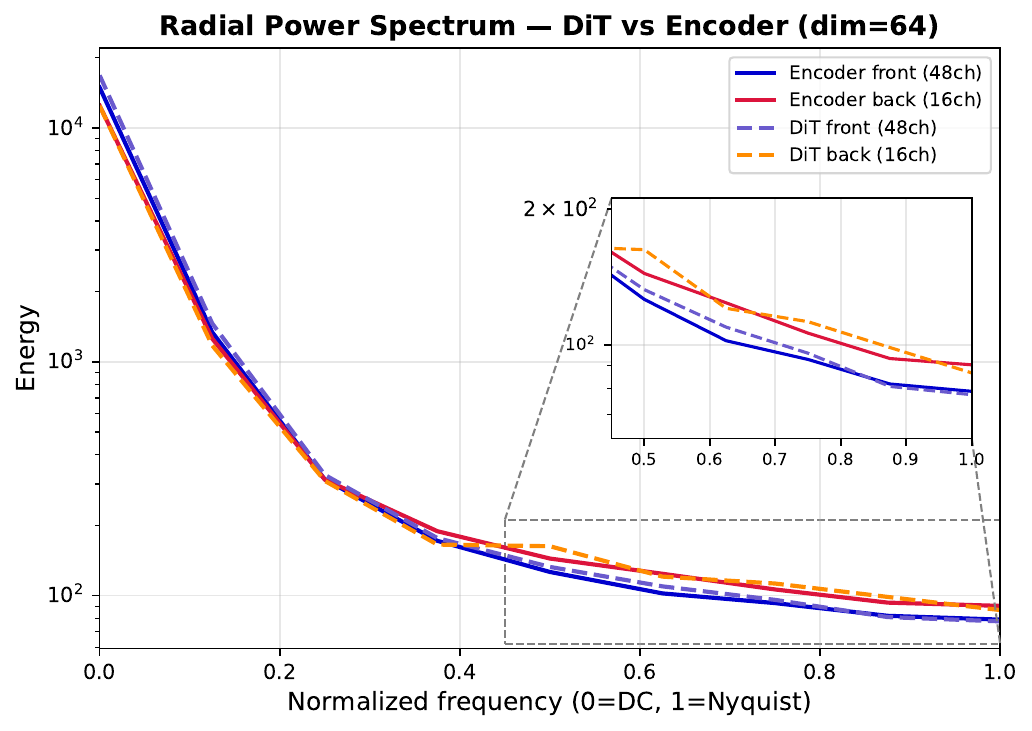}
        \caption{Channel split with $w=16$.}
        \label{fig:appendix_ddt_encoder_b}
    \end{subfigure}

    \vspace{0.8em}

    \begin{subfigure}[t]{0.47\textwidth}
        \centering
        \includegraphics[width=\linewidth]{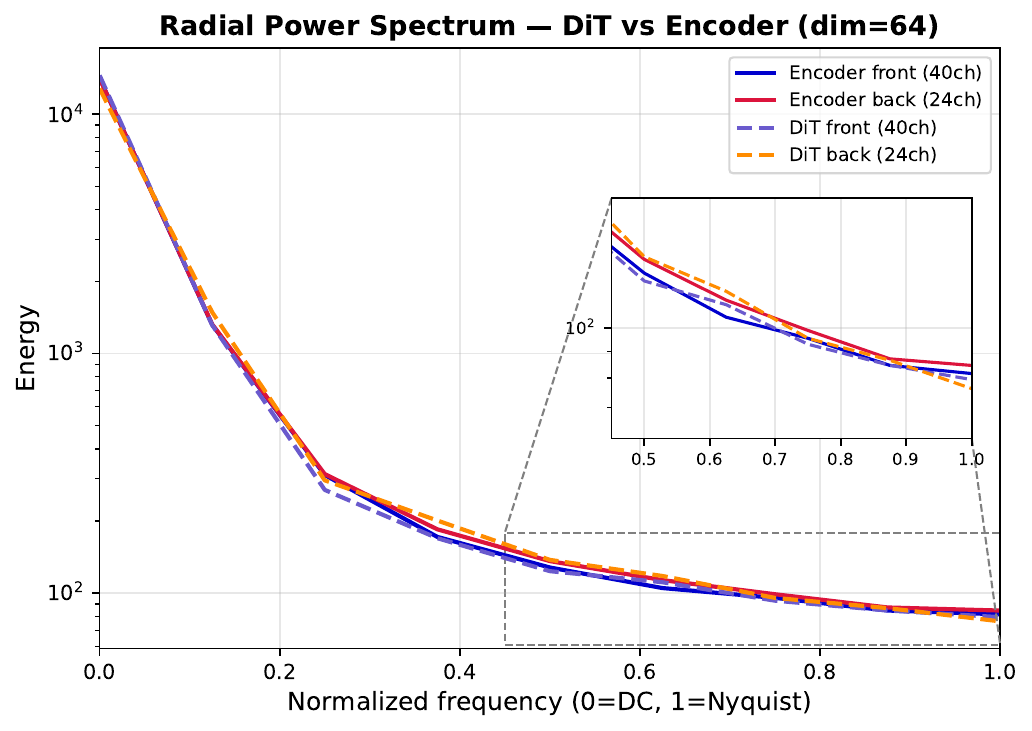}
        \caption{Channel split with $w=24$.}
        \label{fig:appendix_ddt_encoder_c}
    \end{subfigure}
    \hfill
    \begin{subfigure}[t]{0.47\textwidth}
        \centering
        \includegraphics[width=\linewidth]{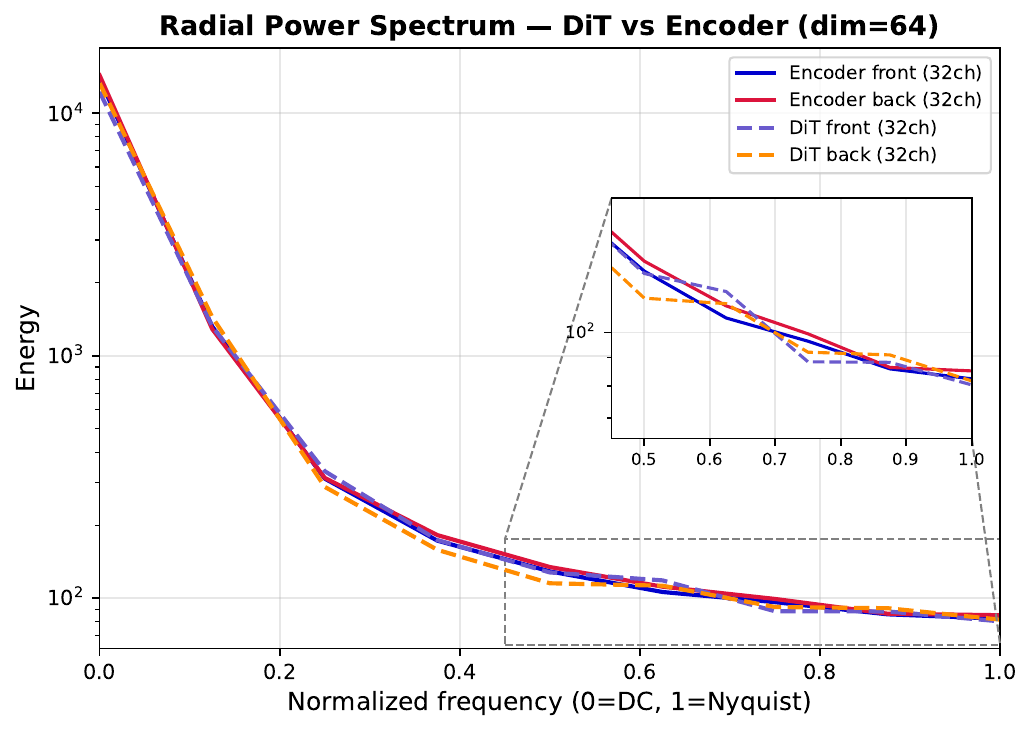}
        \caption{Channel split with $w=32$.}
        \label{fig:appendix_ddt_encoder_d}
    \end{subfigure}
    \caption{Comparison of the radial power spectra of encoder latents and DiT-generated latents across different channel splits in the 64-dimensional bottleneck space.}
    \label{fig:appendix_ddt_vs_encoder_channelwise}
\end{figure*}

\paragraph{Additional spectral analysis.}
We further analyze the spectral alignment between encoder latents and DiT-generated latents under different channel splits in the 64-dimensional bottleneck space. 
For each split, we compare the radial power spectra of encoder latents and generated latents within the corresponding channel groups. 
As shown in Fig.~\ref{fig:appendix_ddt_vs_encoder_channelwise}, earlier channel groups exhibit smoother and more stable spectral behavior, whereas later channel groups contain relatively stronger high-frequency components and are more difficult for the generator to match accurately. 
This trend is consistent with our main observation that structured channel-wise masking induces an ordered latent organization: the front channels primarily capture coarse semantic information, while the back channels encode finer visual details.

\end{document}